\documentclass[conference]{IEEEtran}

\makeatletter

\def\ps@IEEEtitlepagestyle{%
  \def\@oddfoot{\mycopyrightnotice}%
  \def\@evenfoot{}%
}
\def\mycopyrightnotice{%
  {\footnotesize XXX-X-XXXX-XXXX-X/XX/\$XX.00~\copyright~20XX IEEE\hfill}
  \gdef\mycopyrightnotice{}
}

\usepackage{blindtext}
\usepackage{eso-pic}
\IEEEoverridecommandlockouts
\usepackage{cite}
\usepackage{amsmath,amssymb,amsfonts}
\usepackage{algorithmic}
\usepackage{graphicx}
\usepackage{textcomp}
\usepackage{xcolor}
\def\BibTeX{{\rm B\kern-.05em{\sc i\kern-.025em b}\kern-.08em
    T\kern-.1667em\lower.7ex\hbox{E}\kern-.125emX}}
    
\usepackage{eso-pic}
\newcommand\AtPageUpperMyright[1]{\AtPageUpperLeft{%
 \put(\LenToUnit{0.17\paperwidth},\LenToUnit{-2cm}){%
     \parbox{0.9\textwidth}{\raggedleft\fontsize{8}{11}\selectfont #1}}%
 }}%
\newcommand{\conf}[1]{%
\AddToShipoutPictureBG*{%
\AtPageUpperMyright{#1}
}
}    

\usepackage{graphicx}
\usepackage[labelfont=bf]{caption}
\usepackage{url}
\usepackage{float}
\usepackage{hyperref}
\usepackage{amsmath}
 
\usepackage{mathtools}
\usepackage{bm, csquotes}
\usepackage{subcaption}
\usepackage{chngcntr}
\counterwithin{figure}{section}
\usepackage{multirow}
\usepackage{array}
\usepackage{enumerate}
\usepackage{tikz}
\usetikzlibrary{arrows.meta, positioning, shapes.geometric}
\usepackage{booktabs}

\usepackage{todonotes}	

\begin{document}
\title{\vspace*{1cm} Anomaly Detection in Soil Heavy Metal Contamination Using Unsupervised Learning for Environmental Risk Assessment\\
}

\author{\IEEEauthorblockN{Isaac Tettey Adjokatse}
\IEEEauthorblockA{Institute of Environment and Sanitation Studies\\University of Ghana
Legon, Accra, Ghana\\
adjokatseisaac@gmail.com}
\and 
\IEEEauthorblockN{Samuel Senyo Koranteng}
\IEEEauthorblockA{Institute of Environment and Sanitation Studies\\University of Ghana
Legon, Accra, Ghana\\
skoranteng@ug.edu.gh}
\and 
\IEEEauthorblockN{George Yamoah Afrifa}
\IEEEauthorblockA{Ghana Space Science and\\ Technology Institute\\
Accra, Ghana\\
george.afrifa@gaec.gov.gh}
\and
\IEEEauthorblockN{Theophilus Ansah-Narh}
\IEEEauthorblockA{Ghana Space Science and\\ Technology Institute\\
Accra, Ghana\\
theophilus.ansah-narh@gaec.gov.gh}
\and
\IEEEauthorblockN{Marcellin Atemkeng\textsuperscript{$\dagger$}}
\IEEEauthorblockA{Department of Mathematics\\Rhodes University, Grahamstown, South Africa\\
m.atemkeng@ru.ac.za}
\and
\IEEEauthorblockN{Joseph Bremang Tandoh}
\IEEEauthorblockA{Ghana Space Science and\\ Technology Institute\\
Accra, Ghana\\
joseph.tandoh@gaec.gov.gh}
\and
\IEEEauthorblockN{Kow Ahor Essel-Yorke}
\IEEEauthorblockA{Ghana Space Science and\\ Technology Institute\\
Accra, Ghana\\
kow.essel-yorke@gaec.gov.gh}
\and
\IEEEauthorblockN{
Richmond Opoku-Sarkodie}
\IEEEauthorblockA{Department of Information Technology and \\ Mathematical Sciences\\Methodist University 
Ghana, Accra, Ghana\\
ropokusarkodie@gmail.com}
\and
\IEEEauthorblockN{Rebecca Davis}
\IEEEauthorblockA{Department of Actuarial Science\\Pentecost University
 Accra, Ghana\\
rdavis@pentvars.edu.gh}}

\maketitle
\conf{\textit{  Proc. of the International Conference on Electrical, Computer and Energy Technologies (ICECET) \\ 
06-09 July 2026, Rome-Italy}}

\begin{abstract}
Soil contamination by heavy metals poses a persistent environmental and public health concern in rapidly urbanising regions of Ghana, particularly at unregulated waste disposal sites. This study applies an unsupervised machine learning framework to detect and characterise anomalous heavy metal contamination patterns in soils from twelve waste sites and residential controls in the Central Region, of Ghana. Concentrations of eight metals (As, Cd, Cr, Cu, Hg, Ni, Pb, Zn) were analysed alongside standard health risk indices, including the Hazard Index (HI) and Incremental Lifetime Cancer Risk (ILCR).
Isolation Forest and PCA reconstruction error each identified $12$ anomalous samples ($15.4\%$ of $78$ samples), while DBSCAN detected no density-isolated noise points. A consensus approach isolated six robust anomalies ($7.7\%)$, all spatially concentrated at a single site (S3). Anomalies exhibited approximately $70$--$80\%$ higher mean HI values than normal samples, with all consensus anomalies exceeding the HI$=1$ threshold. PCA reconstruction error showed a strong positive association with HI ($r \approx 0.8$), indicating consistency between multivariate deviation and health risk. Three distinct anomaly types were identified: extreme Cu enrichment at S3, anomalously low Ni at S4/S5, and moderate multi-metal (Pb--Zn) co-elevation at S9--S12.
The results demonstrate that unsupervised machine learning provides granular, objective insight beyond aggregate indices, enabling targeted site prioritisation and risk-informed environmental management.
\end{abstract}



\section{Introduction} \label{intro}

Soil contamination by heavy metals represents a persistent and escalating environmental challenge with profound implications for ecosystem integrity and public health.
This issue is particularly acute in regions experiencing rapid industrialisation and inadequate waste management infrastructure. In Ghana, anthropogenic activities, including artisanal mining, vehicle repair, and the informal disposal of electronic and municipal waste, have been identified as significant sources of heavy metal pollution, elevating ecological risks and threatening community health \cite{Kala2025, Mensah2025}.
Studies from areas like Wa in north-western Ghana have documented substantial anthropogenic enrichment of soils with metals such as lead, zinc, and copper, directly linking contamination to local industrial activities. Furthermore, research on abandoned mining sites reveals alarming concentrations of mercury, cadmium, and arsenic in soils and water, leading to calculated Hazard Index (HI) and Incremental Lifetime Cancer Risk (ILCR) values that indicate significant non-carcinogenic and carcinogenic threats to local populations, especially children. These findings underscore the critical need for effective and timely environmental monitoring in vulnerable areas, such as the waste dumping sites in Ghana's Central Region, which are the focus of the present investigation.

Traditional approaches to environmental risk assessment, such as the calculation of contamination indices (e.g., contamination factor, geo-accumulation index), ecological risk indices, and human health risk metrics (HI and ILCR), provide a vital framework for evaluating pollution \cite{fairbrother2007framework}.
However, these methods possess inherent limitations. Primarily, they are aggregate measures that can obscure specific, multi-dimensional contamination signatures. A site may exhibit a moderate overall hazard index while harbouring an extreme and unusual concentration of a single, highly toxic element, a pattern that aggregate scores may dilute. Secondly, these indices often rely on point-in-time sampling and laboratory analysis, which can be resource-intensive, slow, and limited in spatial coverage, hindering the dynamic tracking of pollution spread or the identification of new, atypical contamination sources \cite{Binetti2025}.
Finally, a significant statistical challenge arises from the multicollinearity frequently observed among multiple heavy metal concentrations; their inter-correlations can complicate the interpretation of their individual contributions to risk in multivariate models \cite{Mamouei2022}. This necessitates analytical techniques that can handle such complex, correlated data structures without losing critical information about unique, anomalous events.

The emerging field of machine learning offers powerful, data-driven tools to complement and enhance traditional environmental monitoring. Unsupervised learning algorithms, which require no pre-labelled data on contamination events, are exceptionally well-suited for the exploratory analysis of geochemical datasets to identify anomalous samples that deviate from established background patterns. Techniques such as the Isolation Forest algorithm excel in efficiently isolating anomalies in high-dimensional data by exploiting the principle that anomalous points are \enquote{few and different}, and thus require fewer random partitions to be isolated in a computational tree structure. Similarly, Density-Based Spatial Clustering of Applications with Noise (DBSCAN) is highly effective for identifying clusters of similar data points while flagging outliers that do not conform to the density of any cluster, making it robust for detecting irregular contamination patterns \cite{Proshad2025, Jibrin2025}.
Furthermore, Principal Component Analysis (PCA) serves a dual purpose: it mitigates multicollinearity by transforming correlated variables into a set of uncorrelated principal components, and samples with high reconstruction error from a reduced-component model can themselves be indicative of anomalous, atypical metal assemblages \cite{Vu2019}.
Recent studies have successfully integrated these methods, demonstrating, for instance, that preprocessing soil data with DBSCAN to remove outliers can significantly improve the predictive accuracy of machine learning models for heavy metal concentrations.

Building upon this technological opportunity and addressing the identified gaps in traditional assessment, this study aims to implement a comprehensive unsupervised learning framework to analyse soil heavy metal contamination at selected waste dumping sites in the Central Region of Ghana. The primary objectives are threefold. First, to apply and compare the efficacy of multiple anomaly detection algorithms specifically Isolation Forest, DBSCAN, and PCA-based reconstruction error on a geochemical dataset profiling eight heavy metals--arsenic (As), cadmium (Cd), chromium (Cr), copper (Cu), mercury (Hg), nickel (Ni), lead (Pb), and zinc (Zn).
Second, to identify and characterise soil samples and specific sites that exhibit unusual multi-elemental signatures, which may indicate distinct pollution sources or atypical waste composition not immediately apparent from aggregated hazard indices alone. Third, to validate the anomalies identified by the machine learning consensus against classical risk assessment metrics (HI and ILCR) and site location data, thereby bridging innovative data science techniques with established environmental health paradigms. The ultimate goal is to develop a reproducible analytical framework that can prioritise locations for detailed forensic investigation and inform targeted remediation strategies, contributing to more efficient and proactive environmental management in the region.

\section{Study Area and Data Description} \label{sec:EDA}
\subsection{Study Area}

\begin{figure}
	\begin{minipage}{\linewidth}
		\centering
		\includegraphics[width=\textwidth]{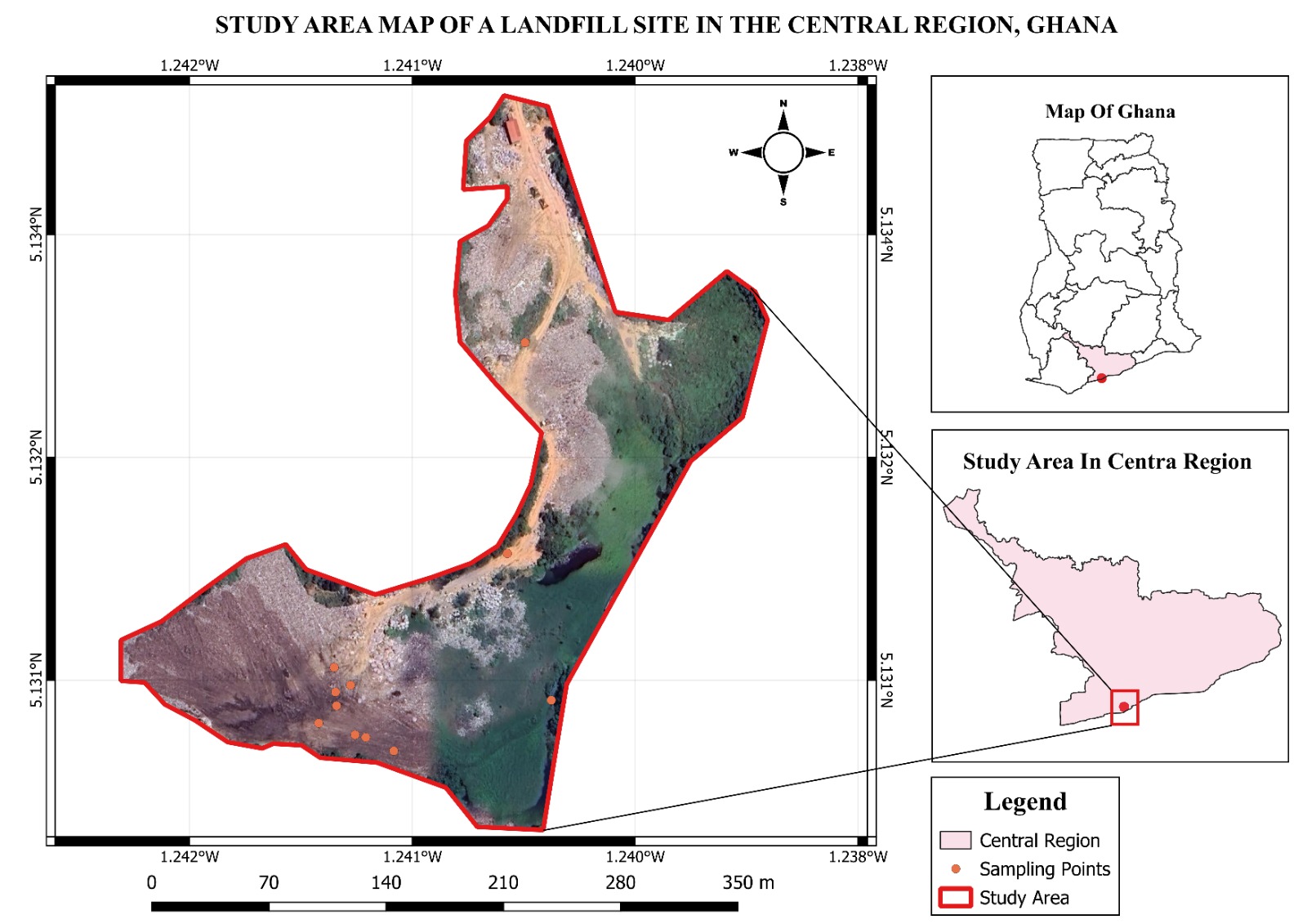} %
	\end{minipage}
	\caption{Study area map showing the locations of the waste dumping sites (S1–-S12).}
	\label{fig:studyAr}
\end{figure}
At each of the twelve waste sites (refer to Fig.~\ref{fig:studyAr}), six discrete topsoil samples (0--15 cm depth) were collected, following established protocols for geochemical surveys \cite{wang2023evaluation}. This sampling strategy was designed to capture within-site variability. All samples were air-dried, sieved to a fine consistency, and subjected to acid digestion. Concentrations of eight priority heavy metals (As, Cd, Cr, Cu, Hg, Ni, Pb, Zn) were determined using Inductively Coupled Plasma Mass Spectrometry (ICP-MS), a method known for its high sensitivity and accuracy for multi-element analysis \cite{wang2024modified}.
For each sample, a human health risk assessment was performed, calculating the non-carcinogenic HI and the ILCR for both adults and children, following standard US Environmental Protection Agency (USEPA) formulas \cite{cobbina2025impact}.
The final dataset comprises geochemical concentrations for the eight metals and their derived risk indices for each individual sample.

This study focuses on the Central Region of Ghana, an area experiencing significant environmental pressure from unregulated waste disposal. To assess the resultant heavy metal contamination, soil samples were systematically collected from twelve distinct waste dumping sites, anonymised as S1 through S12. For comparative baseline assessment, additional control samples were gathered from nearby residential areas with no immediate, direct waste disposal activities, providing a reference for natural background or diffuse urban contamination levels.

Prior to analysis, the dataset underwent a structured preprocessing pipeline. The eight heavy metal concentrations were selected as the core feature set for anomaly detection. To ensure all variables contributed equally to the distance-based and variance-based algorithms and to mitigate the influence of differing measurement scales, the feature set was standardized using the \texttt{StandardScaler} method, which transforms data to have a mean of zero and a standard deviation of one. The risk indices (HI, ILCR) were retained for subsequent validation but were not included in the standardization or the anomaly detection models to prevent circular reasoning. Essential metadata, including the site identifier (S1-S12 or Residential) and a binary flag for control samples, was preserved to enable spatial interpretation and validation of the model outputs.

\section{Anomaly Detection Techniques} \label{sec:SE}

The core analytical framework employs three distinct unsupervised machine learning algorithms to identify anomalous soil samples, each based on a different detection principle. A consensus approach is then used to synthesise their results.

\subsection{Isolation Forest}

The Isolation Forest algorithm is an efficient, tree ensemble method based on the principle that anomalies are \enquote{few and different} and thus easier to isolate from the majority of data points.
It operates by randomly selecting a feature and a split value, recursively partitioning the data. Samples that require fewer partitions to be isolated are deemed more anomalous, as their feature values lie in sparse regions of the data space.
The algorithm outputs an anomaly score for each sample; a lower score indicates a higher degree of anomaly. 
For this study, the model was implemented using the IsolationForest class in \texttt{Python's scikit-learn} library, configured with \texttt{n\_estimators$=200$} (number of trees) and a contamination parameter of $0.15$, assuming a modest proportion of outliers in the dataset.

\subsection{DBSCAN}
 
The DBSCAN identifies anomalies as points in low-density regions that do not belong to any dense cluster.
It defines clusters based on two parameters: \texttt{eps}, the maximum distance between two samples for one to be considered in the neighbourhood of the other, and \texttt{min\_samples}, the minimum number of points required to form a dense region. Points not fulfilling these criteria are labelled as noise (anomalies). The optimal \texttt{eps} value was determined empirically by analysing the $k$-distance plot for the standardized data. The model was implemented with \texttt{min\_samples=$5$}, and samples receiving a cluster label of $-1$ were classified as anomalies.

\subsection{PCA Reconstruction Error} \label{sec:pca}
The PCA is a dimensionality reduction technique that transforms correlated variables into a set of uncorrelated principal components (PCs). In this method, PCA is used to project the standardized $8$-dimensional metal data onto its first two principal components (capturing the major variance structure). Each sample is then reconstructed back to the original $8$-dimensional space from these two PCs. The reconstruction error the difference between the original and reconstructed data is calculated as the Euclidean distance. Samples with a high reconstruction error are poorly represented by the dominant variance patterns of the \enquote{normal} data and are thus flagged as anomalies \cite{boruuvka2005principal}.
A threshold was set at the $85^{\text{th}}$  percentile of the error distribution to label the top $15\%$ of samples as anomalous.

\begin{figure}[H]
\centering
\begin{tikzpicture}[
    node distance=0.5cm,
    every node/.style={font=\small},
    box/.style={rectangle, draw, rounded corners, align=center, minimum width=3.8cm, minimum height=1.1cm},
    bigbox/.style={rectangle, draw, rounded corners, align=center, minimum width=5.2cm, minimum height=1.2cm},
    diamond/.style={diamond, draw, align=center, aspect=2, inner sep=1pt},
    arrow/.style={->, thick}
]

\node[box] (data) {%
\textbf{Data Collection \& Preparation}\\
Soil samples from 12 waste sites (S1-S12)\\
and residential controls in Central Region, Ghana\\
Analysis of 8 heavy metals (As, Cd, Cr, Cu, Hg, Ni, Pb, Zn)\\
Calculation of HI \& ILCR risk indices
};

\node[bigbox, below=of data] (preprocess) {%
\textbf{Data Preprocessing}\\
Feature selection: 8 heavy metal concentrations\\
Standardization of features (StandardScaler)\\
Separation of metadata (Site ID, Control flag)
};

\node[bigbox, below=of preprocess] (detection) {%
\textbf{Parallel Anomaly Detection}\\
\textbf{Isolation Forest}: Contamination=0.15\\
\textbf{DBSCAN}: Density-based clustering (noise as anomaly)\\
\textbf{PCA Recon. Error}: Threshold $= 85^{\text{th}}$ percentile
};

\node[bigbox, below=of detection] (consensus) {%
\textbf{Consensus Anomaly Identification}\\
Voting system: Flag sample if $\geq 2$ methods detect it\\
Output: Final set of consensus anomalies
};

\node[bigbox, below=of consensus] (validation) {%
\textbf{Multi-Factor Validation}\\
Comparison with risk indices (HI $> 1$, ILCR thresholds)\\
Site-wise spatial analysis of anomaly distribution\\
Performance check against residential control samples
};

\draw[arrow] (data) -- (preprocess);
\draw[arrow] (preprocess) -- (detection);
\draw[arrow] (detection) -- (consensus);
\draw[arrow] (consensus) -- (validation);

\end{tikzpicture}
\caption{Schematic overview of the analytical workflow for unsupervised anomaly detection in soil heavy metal contamination. The pipeline proceeds from field data collection and risk index calculation, through standardized preprocessing and parallel application of three detection algorithms, to a consensus-based identification of anomalies, which are finally validated against independent risk metrics and spatial information.}
\label{fig:workflow}
\end{figure}
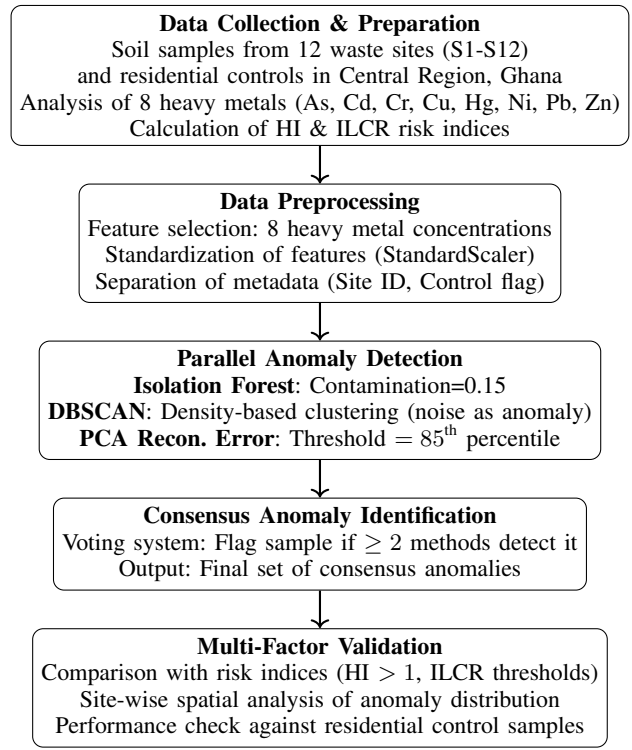

\subsection{Consensus Approach}

To increase the robustness and confidence of anomaly detection, a consensus strategy was adopted. A sample is flagged as a consensus anomaly if it is identified as anomalous by at least two of the three independent methods described above. This voting mechanism mitigates the biases and limitations inherent in any single algorithm, reducing false positives and highlighting the most compelling outliers for further investigation.

The anomalies identified through the machine learning pipeline were validated using multiple lines of evidence. First, the anomaly labels were compared against the computed risk indices. It was hypothesised that consensus anomalies would exhibit significantly higher HI and ILCR values compared to non-anomalous samples. Second, a site-wise analysis was conducted to determine if anomalies were clustered at specific locations (e.g., S3, S4/S5), which would point to localised contamination sources rather than random statistical artefacts. Finally, the residential control samples served as a critical validation set; the expectation was that these samples, representing background conditions, would overwhelmingly be classified as "normal" by the models. This multi-faceted validation grounds the computational findings in established environmental risk metrics and physical reality.

The complete analytical workflow, from raw data to validated anomalies, is visually summarised in the schematic diagram in Fig.~\ref{fig:workflow}.

\section{Results and Discussion} \label{sec:R4}

\subsection{Exploratory Data Analysis}

Table~\ref{tab:metal_statistics} summarises the concentration statistics of the eight analysed heavy metals, revealing moderate enrichment overall but pronounced spatial heterogeneity. This variability is most evident for Cu and Zn, whose large standard deviations relative to their means indicate strong site-specific influences, whereas Hg and Cd exhibit comparatively constrained ranges.

\begin{table*}[htbp]
\centering
\caption{Descriptive statistics of heavy metal concentrations (mg/kg) in soil samples from waste dumping sites in Central Region, Ghana.}
\label{tab:metal_statistics}
\begin{tabular}{l *{8}{r}}
\toprule
\textbf{Statistic} & \textbf{As} & \textbf{Cd} & \textbf{Cr} & \textbf{Cu} & \textbf{Hg} & \textbf{Ni} & \textbf{Pb} & \textbf{Zn} \\
\midrule
Count & 78.00 & 78.00 & 78.00 & 78.00 & 78.00 & 78.00 & 78.00 & 78.00 \\
Mean & 6.48 & 3.35 & 83.88 & 108.65 & 2.44 & 25.03 & 25.65 & 67.43 \\
Std Dev & 3.33 & 1.17 & 24.90 & 148.10 & 0.92 & 14.46 & 19.27 & 44.17 \\
Minimum & 0.01 & 0.01 & 0.01 & 2.35 & 0.01 & 0.12 & 0.03 & 9.34 \\
25th Percentile & 4.75 & 3.24 & 85.54 & 77.47 & 2.14 & 23.70 & 8.45 & 31.64 \\
Median (50th) & 5.20 & 3.53 & 88.65 & 78.66 & 2.54 & 25.46 & 23.75 & 56.00 \\
75th Percentile & 10.48 & 3.70 & 96.42 & 79.85 & 3.15 & 25.73 & 42.33 & 96.05 \\
Maximum & 11.40 & 6.12 & 98.56 & 611.76 & 3.78 & 52.70 & 56.44 & 165.78 \\
\bottomrule
\end{tabular}
\end{table*}

\begin{figure*}
	\begin{minipage}{\linewidth}
		\centering
		\includegraphics[width=\textwidth]{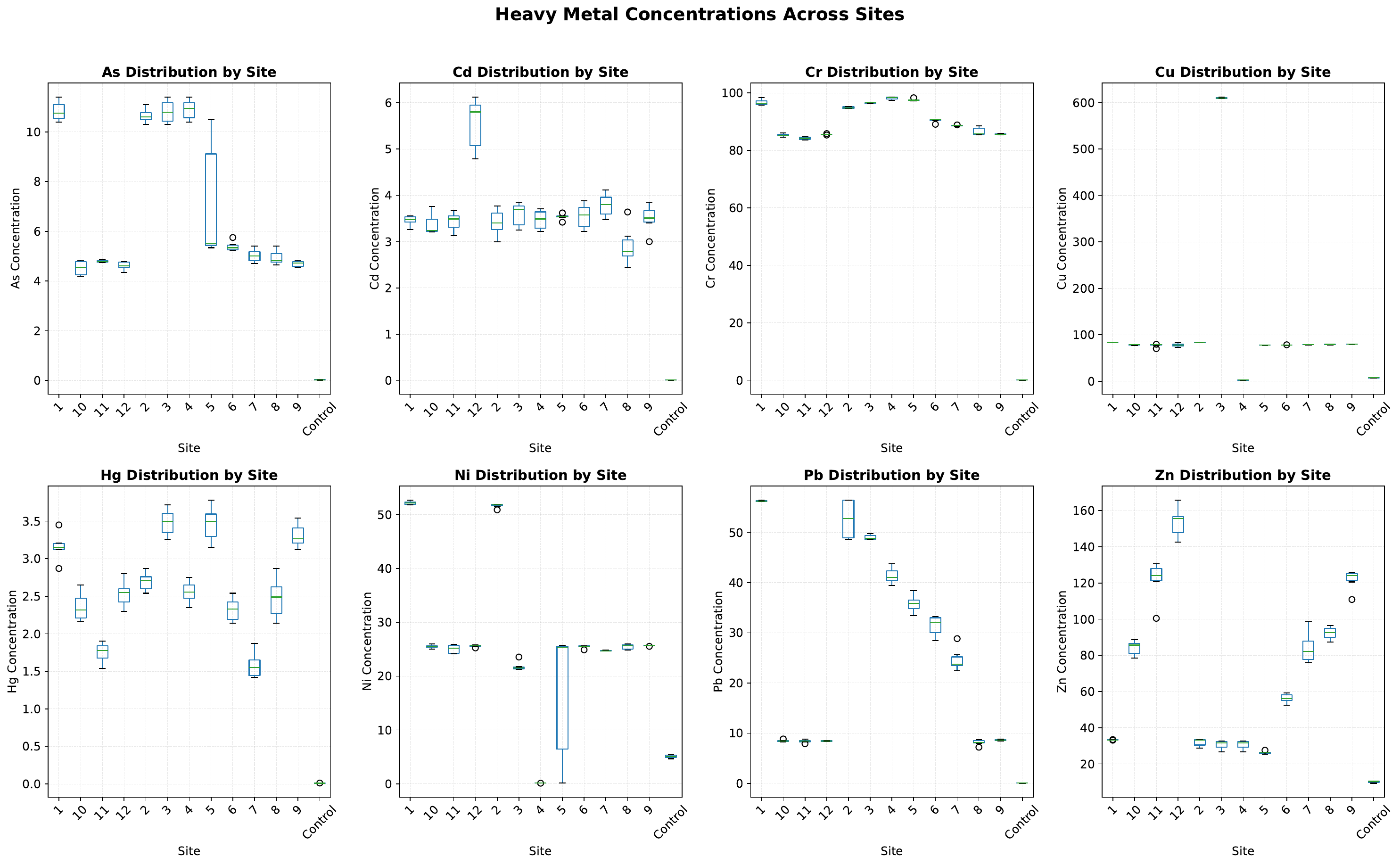} %
	\end{minipage}
	\caption{Site-wise boxplots of As, Cd, Cr, Cu, Hg, Ni, Pb, and Zn concentrations for waste and control locations.}
	\label{fig:FI_metal_distributions_by_site}
\end{figure*}

Figure~\ref{fig:FI_metal_distributions_by_site} illustrates the site-wise distributions, highlighting an extreme Cu outlier at Site~S3 ($\sim$612~mg/kg), which dominates the overall Cu variability and suggests a localised source contribution. In contrast, Ni concentrations at Sites~S4 and S5 are consistently low, indicating site-dependent geochemical or waste-related controls rather than random variability. Other metals show moderate dispersion across sites, with As and Cr displaying relatively uniform behaviour, while control samples generally occupy the lower concentration ranges.

\begin{figure*}
	\begin{minipage}{\linewidth}
		\centering
		\includegraphics[width=0.6\textwidth]{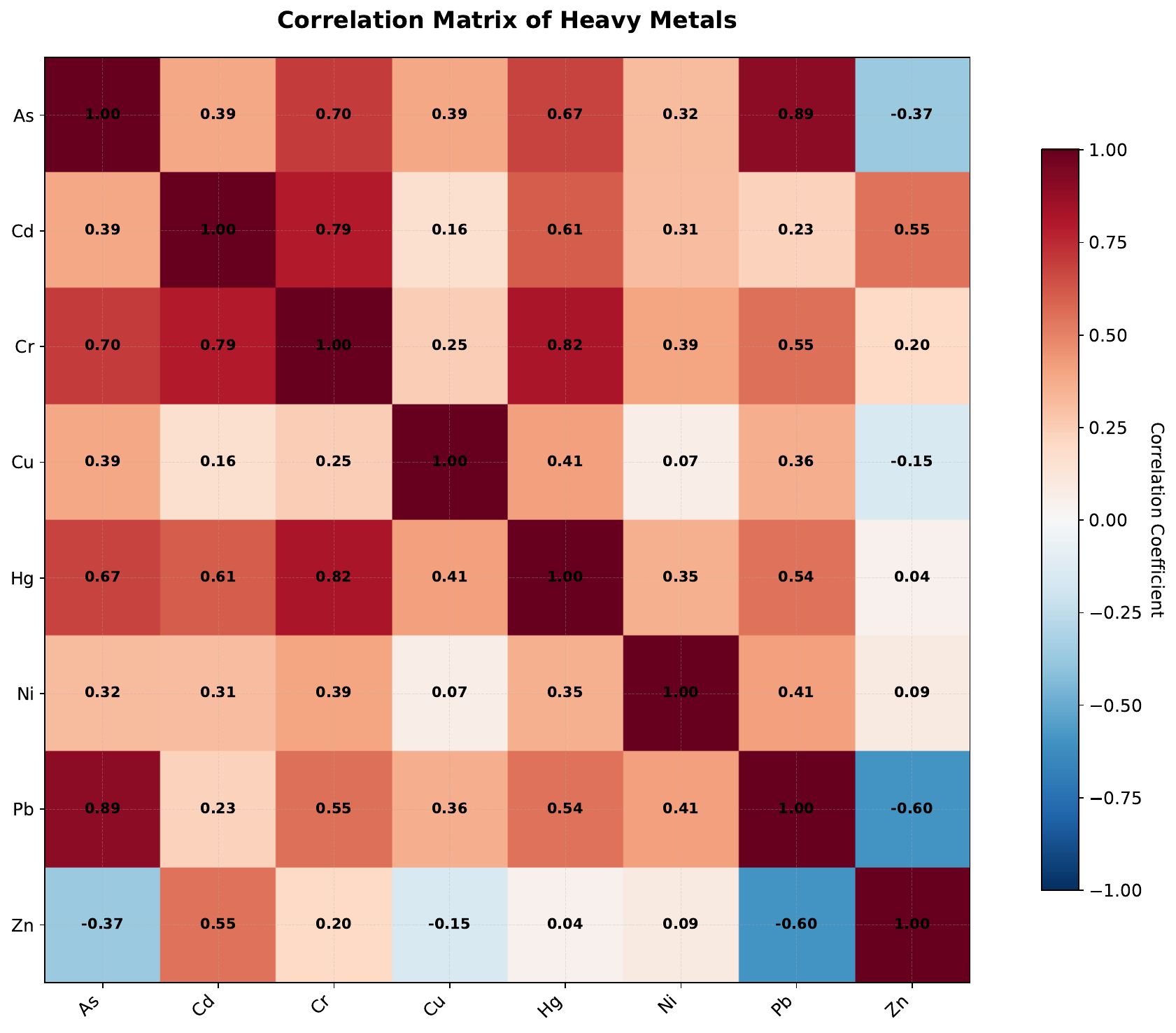} %
	\end{minipage}
	\caption{Pearson correlation matrix of heavy metal concentrations across all samples.}
	\label{fig:FII_correlation_heatmap}
\end{figure*}

The Pearson correlation matrix in Fig.~\ref{fig:FII_correlation_heatmap} reveals several notable inter-metal associations, including strong positive correlations for Cr--Hg ($r=0.82$), Cd--Cr ($r=0.79$), and As--Pb ($r=0.89$), consistent with partial co-occurrence from mixed waste inputs. Zn shows negative correlations with As and Pb, suggesting divergent source influences or retention mechanisms, while Cu exhibits weaker correlations overall, reinforcing its site-specific anomalous behaviour.

Collectively, these results demonstrate marked spatial heterogeneity and a mixed correlation structure, supporting the need for multivariate and unsupervised anomaly detection approaches beyond univariate or aggregated indices.

\subsection{Anomaly Detection Results}


\begin{figure}
	\begin{minipage}{\linewidth}
		\centering
		\includegraphics[width=1.0\textwidth]{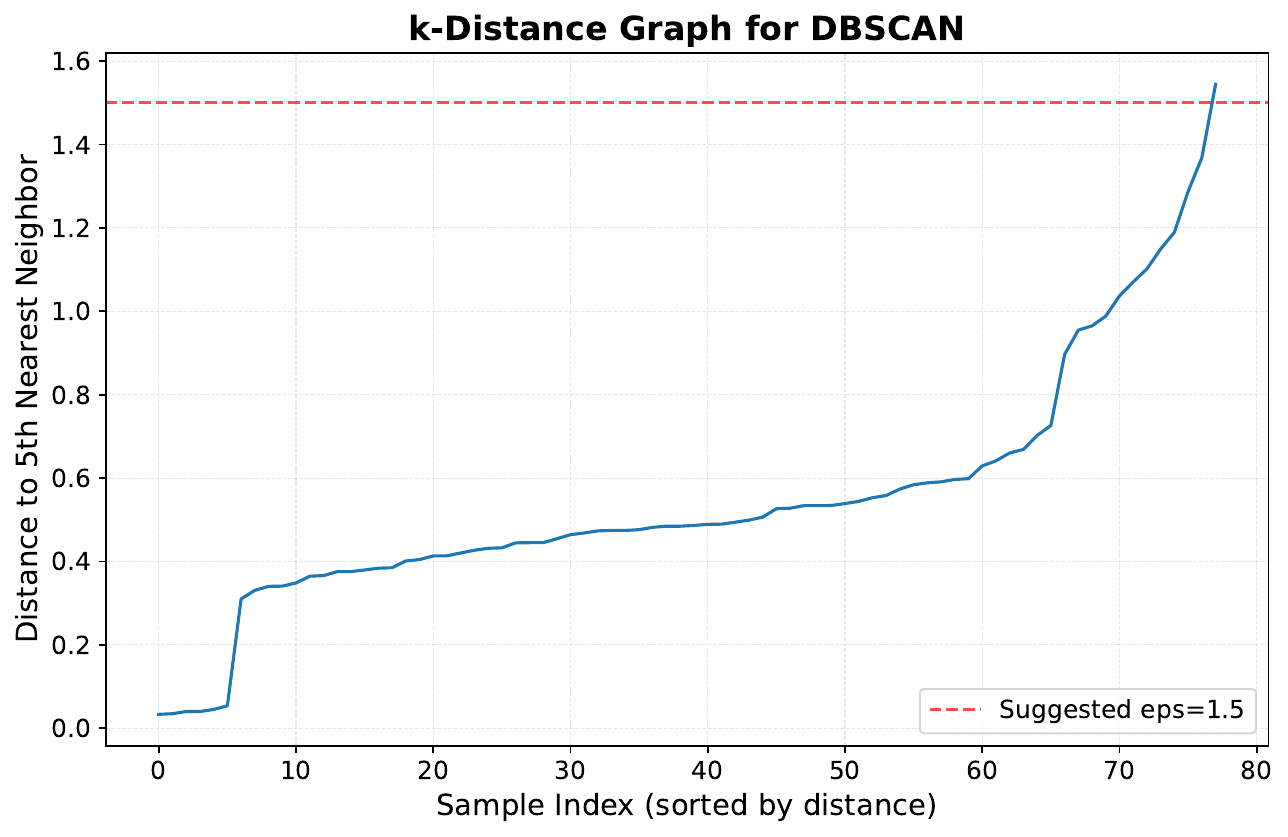} %
	\end{minipage}
	\caption{$k$-distance plot for DBSCAN parameter selection, showing the inflection point used to define the neighbourhood radius.
}
	\label{fig:FIV_dbscan_k_distance}
\end{figure}

The $k$-distance graph in Fig.~\ref{fig:FIV_dbscan_k_distance} exhibits a clear inflection point near $\varepsilon=1.5$, which was adopted as the DBSCAN neighbourhood radius. This choice balances cluster compactness and noise separation, enabling robust identification of samples in low-density regions without excessive fragmentation of the main data structure.

\begin{figure*}
	\begin{minipage}{\linewidth}
		\centering
		\includegraphics[width=1.0\textwidth]{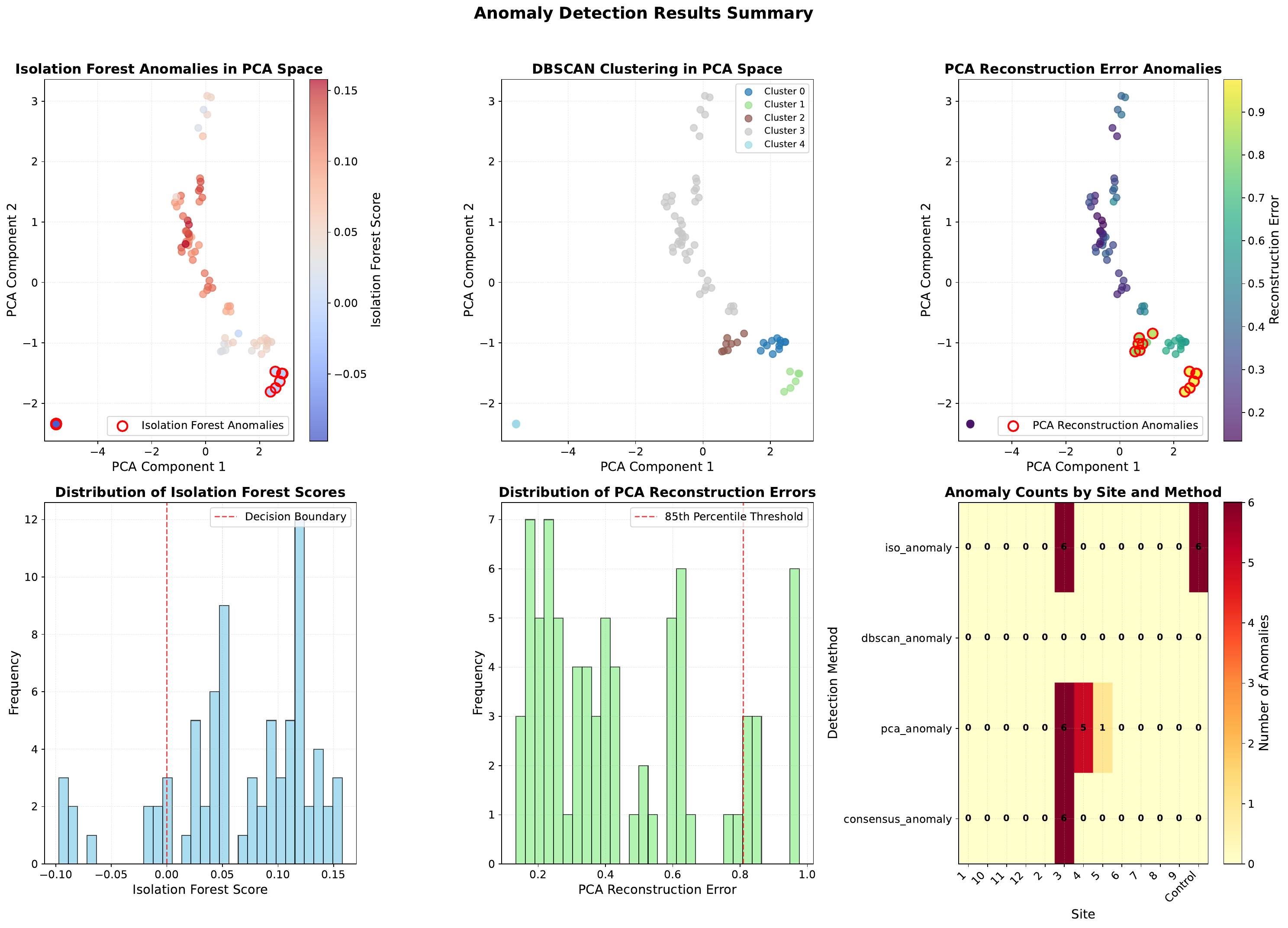} %
	\end{minipage}
	\caption{PCA space visualisation with anomaly overlays from Isolation Forest, DBSCAN, and PCA reconstruction error, incorporating the anomaly count matrix by site and detection method, and illustrating cross-method agreement and the spatial concentration of consensus anomalies.
}
	\label{fig:FV_anomaly_detection_results}
\end{figure*}


PCA-based visualisations with anomaly overlays (Fig.~\ref{fig:FV_anomaly_detection_results}) reveal coherent groupings in the reduced feature space across all methods, reflecting shared structure in the multivariate metal concentrations. Isolation Forest anomalies occupy sparsely populated regions of this space, including a small number of samples from the residential control group. This behaviour arises from the algorithm’s global partitioning strategy, which is sensitive to marginal multivariate deviations but does not explicitly incorporate local density information.

DBSCAN, as a density-based method, formalises these visual groupings into five compact clusters and assigns no samples to the noise class. Under the selected neighbourhood radius ($\varepsilon=1.5$) and minimum sample criterion, all observations, including controls, are embedded within locally dense regions. The absence of DBSCAN anomalies indicates that the detected outliers are not density-isolated but instead lie within broader concentration gradients, a characteristic commonly observed in heterogeneous environmental datasets.

PCA reconstruction error identifies samples that are poorly represented by the dominant variance structure, with detections concentrated at Site~S3 and partially overlapping with Isolation Forest anomalies. Notably, samples flagged by Isolation Forest within the residential control group are not supported by DBSCAN or PCA reconstruction error and are therefore excluded from the consensus anomaly set.

The anomaly count matrix embedded in Fig.~\ref{fig:FV_anomaly_detection_results} (refer to the matrix table) confirms this filtering effect. Isolation Forest and PCA reconstruction error each flag 12 samples (15.4\% of the 78 analysed samples), while DBSCAN identifies no anomalies. Applying the consensus criterion isolates six samples (7.7\%), all originating from Site~S3, with no consensus anomalies in control locations. This convergence indicates that the consensus anomalies represent site-specific, multivariate deviations associated with elevated HI and ILCR values, rather than artefacts of algorithm-specific sensitivity or background variability.

\section{Conclusion}

This study presents an unsupervised machine learning framework for detecting and characterising anomalies in soil heavy metal contamination, demonstrating its utility as a complementary tool to conventional environmental risk assessment. The framework successfully identified three distinct anomaly types, including extreme single-element enrichment and more subtle multi-metal signatures, each with implications for potential contamination sources. By integrating machine learning outputs with established health risk indices, the study illustrates a practical synergy between data-driven anomaly detection and traditional risk-based evaluation.

The findings provide actionable guidance for environmental management. Site~S3 should be prioritised for detailed investigation and remediation due to extreme Cu enrichment, while Sites~S4 and S5 warrant continued monitoring to clarify the persistence and drivers of anomalous Ni behaviour. The use of a consensus anomaly detection strategy is recommended to enhance robustness and minimise method-specific bias. More broadly, incorporating unsupervised machine learning as an initial screening layer within environmental monitoring programmes can improve the efficiency of site prioritisation under limited resources.

Future work will focus on extending the framework to address current limitations. Incorporating spatial autocorrelation and GIS-based analysis will enable more explicit treatment of spatial dependency, while expansion to larger, multi-regional datasets will improve generalisability. Temporal analysis will allow assessment of contamination dynamics, and integration with real-time data streams from IoT-enabled sensors offers potential for continuous monitoring. Finally, exploring semi-supervised approaches that incorporate expert validation may further enhance interpretability and decision support.

\section*{Acknowledgment}
The authors thank the anonymous reviewers for their constructive comments, which helped improve the clarity and quality of this manuscript. Prof.~Atemkeng also acknowledges financial support from Rhodes University, South Africa, towards the publication of this work.

\bibliographystyle{IEEEtran}
\bibliography{refs}

\end{document}